%% file: paper.tex
\pdfoutput=1

\documentclass[11pt]{article}

 \usepackage[preprint]{acl}

\usepackage{times}
\usepackage{latexsym}

\usepackage[T1]{fontenc}

\usepackage[utf8]{inputenc}

\usepackage{microtype}

\usepackage{inconsolata}

\usepackage{graphicx}

\usepackage{xspace,mfirstuc,tabulary}
\usepackage[section]{algorithm}
\usepackage[noend]{algpseudocode}
\usepackage{relsize}
\usepackage{caption}
\usepackage{subcaption}
\usepackage{amsmath}
\usepackage{bm}
\usepackage{amsfonts}
\usepackage{booktabs}
\usepackage{multirow}
\usepackage{mathtools}
\usepackage{siunitx}
\input{math_commands.tex}


\newcommand{\Tref}[1]{Table~\ref{#1}}

\title{Edit-Constrained Decoding for Sentence Simplification}

\author{Tatsuya Zetsu \\
  LY Corporation \\
  Japan \\
  \texttt{zetsu.t4@gmail.com} \\\And
  Yuki Arase \\
  Tokyo Institute of Technology \\
  Japan \\
  \texttt{arase@c.titech.ac.jp} \\\And
  Tomoyuki Kajiwara \\
  Ehime University \\
  Japan \\
  \texttt{kajiwara@cs.ehime-u.ac.jp} \\
  }

\begin{document}
\maketitle
\begin{abstract}
We propose edit operation based lexically constrained decoding for sentence simplification. 
In sentence simplification, lexical paraphrasing is one of the primary procedures for rewriting complex sentences into simpler correspondences. 
While previous studies have confirmed the efficacy of lexically constrained decoding on this task, their constraints can be loose and may lead to sub-optimal generation. 
We address this problem by designing constraints that replicate the edit operations conducted in simplification and defining stricter satisfaction conditions. 
Our experiments indicate that the proposed method consistently outperforms the previous studies on three English simplification corpora commonly used in this task. 
\end{abstract}

\section{Introduction}
Lexically constrained decoding~\cite{anderson-etal-2017-guided,post-vilar-2018-fast,hu-etal-2019-improved} allows to explicitly apply human knowledge in language generation, which has been employed in various tasks. 
Applications include machine translation using a bilingual dictionary of technical terms as constraints~\cite{chatterjee-etal-2017-guiding,hokamp-liu-2017-lexically}, data augmentation using references as constraints~\cite{geng-etal-2023-improved}, style transfer using style-specific vocabulary as constraints~\cite{kajiwara-2019-negative}, knowledge-grounded generation~\cite{choi-etal-2023-kcts}, and commonsense reasoning using common concepts as constraints~\cite{lu-etal-2021-neurologic}. 
The notable advantage of lexically constrained decoding is its direct applicability to decoders without the need for model (re)training.

Lexically constrained decoding is particularly appealing for sentence simplification, where the transformation of complex tokens to simpler ones is crucial. 
\citet{Laufer1989} points out that learners of a foreign language need to know $95\%$ of tokens in the input text to comprehend the text message. 
A recent evaluation metric for sentence simplification~\cite{heineman-etal-2023-dancing} also focuses on word-level edit quality. 
Previous studies employed negative and positive lexical constraints, i.e., to suppress the output of difficult tokens (negative constraints)~\cite{dehghan-etal-2022-grs} and to encourage the output of their corresponding simpler alternatives (positive constraints)~\cite{zetsu-etal-2022-lexically}. 
These studies showed that such simple constraints significantly improve the quality of simplification. 
However, their naive constraints are not tight enough, which leads to sub-optimal simplification satisfying easier constraints yet failing on difficult ones. 

\begin{figure}[t]
    \centering
    \includegraphics[width=0.98\linewidth]{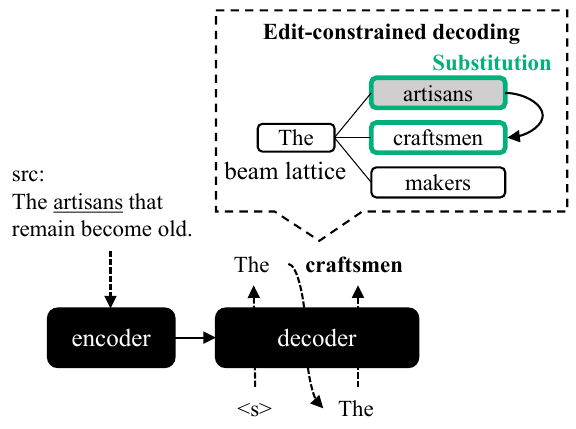}
    \caption{Our method constrains generation based on edit operations during beam search. Here, `artisans' is replaced by `craftsmen' by a substitution constraint.}
    \label{fig:proposal_image}
\end{figure}

To address this problem, we design constraints more directed to simplification associated with stricter satisfaction conditions. 
Sentence simplification conducts three edit operations to rewrite sentences, i.e., \emph{insertion}, \emph{deletion}, and \emph{substitution} of tokens.
We expand NeuroLogic decoding~\cite{lu-etal-2021-neurologic} to perform these edit operations via constrained decoding. 
\Figref{fig:proposal_image} illustrates our method that constrains generation using a substitution operation during beam search. 

We evaluate our method using standard sentence simplification corpora that are publicly available, namely, Turk~\cite{xu-etal-2016-optimizing}, ASSET~\cite{alva-manchego-etal-2020-asset}, and AutoMeTS~\cite{van-etal-2020-automets}. 
The experiment results using oracle constraints extracted from references confirm that the proposed method largely outperforms previous methods utilising lexically constrained decoding. 
Furthermore, the results when constraints are predicted by a simple model indicate the consistent efficacy of the proposed method over these baselines. 
The implementation of our edit-constrained decoding, as well as model outputs, are available at \url{https://github.com/t-zetsu/EditConstrainedDecoding}.

\section{Related Work}
\label{sec:related_work}
\subsection{Sentence Simplification}
Sentence simplification~\cite{shardlow-2014,alva-manchego-etal-2020-data} paraphrases complex sentences into simpler forms. 
The primary approaches can be categorised into three types: (a) translation-based, (b) edit-based, and (c) hybrid approaches.
The translation-based approach, e.g., ~\cite{nisioi-2017,zhang-2017,kriz-2019,surya-2019,sheang-saggion-2021-controllable,martin-etal-2022-muss,anschutz-etal-2023-language}, formalises sentence simplification as monolingual machine translation from complex to simple sentences. 
This approach learns rewriting patterns from corpora and thus allows flexible rewriting; however, the infrequent nature of simplification operations hinders a model from learning necessary operations. 
As a result, it tends to maintain complex tokens intact and end up in too conservative rewriting~\cite{zhao-etal-2018-integrating, kajiwara-2019-negative}. 

In contrast, the edit-based approach~\cite{alva-manchego-etal-2017-learning,dong-2019,kumar-2020,mallinson-2020,omelianchuk-2021} rewrites a source sentence by editing, i.e., deleting, inserting, and replacing, its tokens. 
This approach can address the conservativeness problem owing to explicit token-by-token edits. 
However, it lacks the flexibility to rewrite an entire sentence to change its structure~\cite{zetsu-etal-2022-lexically}. 

Finally, the hybrid approach gets the best of both worlds by applying lexical constraints to translation-based models. 
\citet{agrawal-2021} biases a non-autoregressive simplification model by setting an initial state of decoding, considering the lexical complexity of a source sentence. 
\citet{kajiwara-2019-negative} and \citet{dehghan-etal-2022-grs} apply a negative lexical constraint using lexically constrained decoding to avoid outputting complex tokens while \citet{zetsu-etal-2022-lexically} employ both positive and negative lexical constraints. 
In line with these previous studies, we further enhance the hybrid approach by defining constraints based on edit operations. 

\subsection{Enhanced Constrained Decoding}
Earlier studies on lexically constrained decoding have focused on computational efficiency~\cite{post-vilar-2018-fast,hu-etal-2019-improved,Miao_Zhou_Mou_Yan_Li_2019,sha-2020-gradient}. 
Since its prevalence in text generation tasks, recent studies have expanded the types of constraints considered. 
\citet{bastan-etal-2023-neurostructural} enhanced NeuroLogic decoding to consider syntactic constraints, namely, dependency types up to two entities. 
Similarly, \citet{geng-etal-2023-grammar} proposed constrained decoding by grammar and \citet{mccarthy-etal-2023-long} proposed finite-state constraints. 
Transformation of syntactic structure is another primary factor in simplification~\cite{niklaus-etal-2019-transforming}. 
In our method, structure transformation is implicitly handled by the base translation-based model. 
Our constraints are lexical but expand the simple concepts of positive and negative constraints to replicate edit operations conducted in sentence simplification. 


\section{Edit-Constrained Decoding}
We expand NeuroLogic Decoding~\cite{lu-etal-2021-neurologic} to realise edit-based constraints, for its efficacy and computational efficiency when applied to sentence simplification~\cite{zetsu-etal-2022-lexically}.  

\subsection{Preliminary: NeuroLogic Decoding}
We briefly review NeuroLogic Decoding. 
Conceptually, it seeks for a hypothesis with the highest generation likelihood and constraint satisfaction: 
\begin{equation}\label{eq:neurologic}
    \argmax_{Y \in \mathcal{Y}} P_\theta(Y|X) \text{ s.t. } \sum_{i=1}^L C_i=L,
\end{equation}
where $P_\theta(X|Y)$ is the likelihood to generate a sequence $Y=\{y_0,y_1,\cdots, y_N\}$ given an input sequence $X=\{x_0,x_1,\cdots,x_M\}$, $\mathcal{Y}$ is the possible generation space, and $C_i$ is $i$-th positive or negative constraint (in total $L$ constraints). 
Specifically, NeuroLogic Decoding achieves this by expanding the beam search to conduct \textbf{pruning}, \textbf{grouping}, and \textbf{selecting} processes at every timestep. 

NeuroLogic Decoding first \textbf{prunes} candidates by discarding hypotheses confirmed as non-satisfying constraints and then filtering out less likely hypotheses with fewer constraint satisfaction. 
Namely, it drops any candidates not in the top-$\alpha$ in terms of likelihood and not in the top-$\beta$ in terms of number of satisfied conditions. 
These $\alpha \in \mathbb{N}^+$ and $\beta \in \mathbb{N}^+$ are hyper-parameters. 
Next, it \textbf{groups} the remaining candidates based on the states of constraint satisfaction, and 
then \textbf{selects} a hypothesis from the candidates in each group that preserves high generation likelihood and the larger number of constraint satisfaction. 

These grouping and selecting steps aim to diversify candidates in a beam and broaden the search space to avoid biasing toward hypotheses being highly likely yet satisfying only easy constraints at early timesteps. 
However, as we discuss in \Secref{sec:edit-based-constraints}, further consideration is needed for properly handling negative constraints (i.e., deletion operations). 
For more details on NeuroLogic Decoding, please refer to \citet{lu-etal-2021-neurologic}.

\subsection{Edit-based Constraints}
\label{sec:edit-based-constraints}

\begin{figure*}[t]
    \centering
    \includegraphics[width=0.87\textwidth]{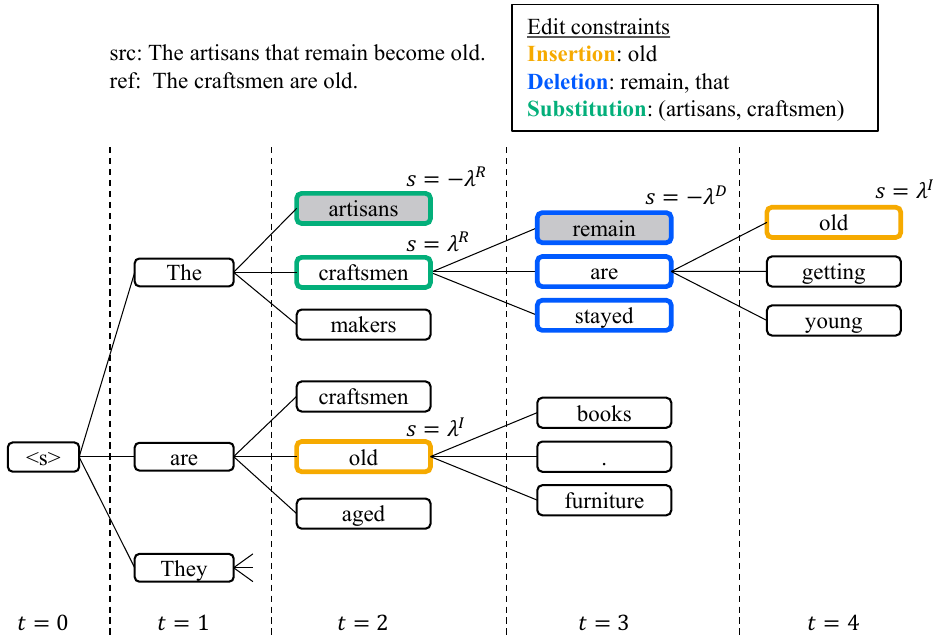}
    \caption{Edit-based constraint decoding example (beam size is $3$)}
    \label{fig:edit-based-constraints}
\end{figure*}

Inspired by the edit operations conducted in sentence simplification, we define edit constraints of \emph{insertion}, denoted as $C^I$, \emph{deletion}, denoted as $C^D$, and \emph{substitution}, denoted as a tuple $C^R=(C^{R_{i}}, C^{R_{o}})$ where $C^{R_{i}}$ is replaced by $C^{R_{o}}$. 
These edit constraints are associated with satisfaction and dissatisfaction conditions. 
Furthermore, instead of a binary weight, we use a numerical score $s$ to allow flexibility in appreciating different types of constraints. 
Therefore, the second term in \Eqref{eq:neurologic} changes to maximise the total edit score: 
\begin{equation}
\label{eq:edit_score}
    \sum_{i=0}^t y_i(s),
\end{equation}
where $y_i(s)$ denotes the score that the $i$-th output token receives and $t$ is the current timestep. 
Note that $y_i$ corresponds to a node $n_j$ in the lattice of beam search; we use $y_i$ and $n_j$ interchangeably in the following for notation simplicity.

\paragraph{Insertion} operation is equivalent to the positive constraint. 
If the node $n_i$ is equivalent to the $j$-th insertion constraint $C^I_j$,\footnote{We use exact matching of node and constraint tokens.} the $n_i$ receives the score $s$ as follows.  
\begin{equation}  \label{eq:insertion}
s=
    \begin{dcases}
        \lambda^I    &   \text{if $n_i=C^I_j$},  \\
        0            &   \text{otherwise},
    \end{dcases}
\end{equation}
where $\lambda^I \in \mathbb{R}^+$ is a weight of insertion operation. 

\paragraph{Deletion} operation conceptually corresponds to the negative constraint. 
Previous studies~\cite{kajiwara-2019-negative,dehghan-etal-2022-grs,zetsu-etal-2022-lexically} naively defined negative constraints, i.e., they regard a negative constraint as satisfied if a certain word is simply avoided, without further conditions. 
However, we argue that the deletion operation, or negative constraint, requires more careful consideration to be properly handled. 
Otherwise, we may end up with a sub-optimal generation that satisfies only negative constraints $C^D$ at early timesteps. 
This is because the size of the vocabulary is typically much larger than that of deletion constraints; their satisfaction is far easier than the positive counterpart. 
As a result, hypotheses that have high likelihoods and exclude negative constraints at early timesteps remain in a beam, which struggles to satisfy harder positive constraints later on. 
For example, in \Figref{fig:edit-based-constraints}, the second beam candidate `are old .' can be the best hypothesis because it satisfies all positive (`old') and negative (exclusion of `remain' and `aged') constraints and has high likelihood due to its shorter length.\footnote{Length normalisation in beam search should alleviate this problem in general; however, we empirically confirmed that adjustment of length normalisation parameter has little effect on our problem.}


To address this problem, we design a stricter condition for the satisfaction of deletion constraint. 
Intuitively, the deletion operation is regarded as satisfied by selecting sibling nodes other than the one equivalent to the constraint. 
Specifically, the node $n_i$ is regarded as satisfying the deletion constraint $C^D_j$ and avoids a penalty (negative score) if and only if its sibling node $n_k \in \pi$ is equivalent to $C^D_j$, where $\pi$ denotes the set of sibling nodes of $n_i$ and $n_k$ in the lattice. 
These nodes receive the score $s$: 
\begin{equation}  \label{eq:deletion}
s=
    \begin{dcases}
        -\lambda^D    &   n_k=C^D_j, \\
        0             &   \forall n_i \in \{\pi \setminus n_k=C^D_j\},
    \end{dcases}
\end{equation}
where $\lambda^D \in \mathbb{R}^+$ is a weight of deletion operation. 
Note that no penalty is given when no node in siblings is equivalent to $C^D_j$, i.e., the score $s=0$.

\paragraph{Substitution} operation replaces a token $C^{R_{i}}$ to $C^{R_{o}}$. 
You may think that this operation can be realised by a combination of the insertion and deletion operations (or positive and negative constraints); however, it would lead to sub-optimal generation due to the lack of a mechanism to ensure the simultaneous satisfaction of both constraints like in~\cite{zetsu-etal-2022-lexically}. 
Therefore, similar to the deletion operation, we design a strict satisfaction condition for substitution. 
The constraint is regarded as satisfied when selecting a node $n_i$ equivalent to $C^{R_{o}}$ if and only if $n_i$ has a sibling node $n_k \in \pi$ equivalent to $C^{R_{i}}$. 
These sibling nodes receive the score $s$: 
\begin{equation}  \label{eq:substitution}
s=
    \begin{dcases}
        -\lambda^R    &  n_k = C^{R_{i}},\\
        \lambda^R     &  n_i = C^{R_{o}}, 
    \end{dcases}
\end{equation}
where $\lambda^R \in \mathbb{R}^+$ is a weight of substitution operation. 
When there is no pair of nodes equivalent to $C^{R_{i}}$ and $C^{R_{o}}$ in $\pi$, these nodes receive no reward nor penalty, i.e., $s=0$. 

\subsection{Algorithm}
Our method conducts the same pruning, grouping, and selecting processes as NeuroLogic Decoding at each timestep of beam search. 
The only difference is the pruning conditions. 
Namely, our method drops candidates whose edit scores (\Eqref{eq:edit_score}) are less than $\delta \in \mathbb{R}^+$ from the top score.\footnote{Remind that NuroLogic Decoding filters out candidates other than top-$\beta$ ones.}  

\Figref{fig:edit-based-constraints} illustrates an example of edit-constrained decoding with a beam size of $3$. 
At timestep $2$, in the first beam, the substitution constraint is satisfied as the token `artisans' is replaced by its sibling of `craftsmen.' 
These nodes receive the penalty and reward scores, respectively. 
In contrast, in the second beam, the substitution constraint is unsatisfied because the token `craftsmen' appears in siblings but `artisans' does not. 
At timestep $3$, the deletion constraint is satisfied in the first beam as `remain' appears among siblings. 
Finally, at timestep $4$, the insertion constraint of `old' is satisfied. 

\section{Experiment Settings}

We conducted two kinds of evaluations. 
The first evaluation (\Secref{sec:oracle_eval}) aims to compare the performance of lexically constrained decoding with different constraint designs. 
To make the effects of methodological differences clearer, this evaluation used oracle constraints extracted from references. 
The second evaluation (\Secref{sec:practical_eval}) follows to observe the performance under a practical setting with predicted constraints. 
Preceding these evaluations, this section describes the common experiment settings.  

\subsection{Datasets} 
We measured the performance on three sentence simplification corpora, namely, \textbf{Turk}~\cite{xu-etal-2016-optimizing}, \textbf{ASSET}~\cite{alva-manchego-etal-2020-asset}, and \textbf{AutoMeTS}~\cite{van-etal-2020-automets}. 
Turk and ASSET are commonly used public evaluation corpora in simplification. 
In addition, we also included AutoMeTS, aiming to investigate the effects of our method in a domain where technical terms are prevalent, and thus, lexical paraphrasing is crucial.  

\paragraph{Turk} focuses on lexical paraphrasing, i.e., replacing difficult tokens with simpler synonyms. 
Turk was created by crowd-sourcing, where annotators were instructed to rewrite sentences by reducing the number of difficult tokens or idioms. 

\paragraph{ASSET} expanded the Turk corpus to encompass a variety of rewriting patterns, not only lexical paraphrasing but also sentence splitting and deleting unimportant information. 
ASSET uses the same source sentences with Turk and crowdsourced these dynamic rewrites. 

\paragraph{AutoMeTS} is a simplification dataset in the medical domain. 
It extracted pairs of expert-oriented and simpler sentences from the sentence-aligned English Wikipedia corpus~\cite{kauchak-2013-improving} using the medical term dictionary selected from the Unified Medical Language System~\cite{umls}. 
This dataset contains technical terms frequently that are expected to be properly rephrased. 
We discarded pairs whose source and reference were identical, which reduced the number of pairs from what was reported in the original paper.

\begin{table}[t]
\centering
\begin{tabular}{@{}cccc@{}}
\toprule
         & Train                      & Validation                      & Test                   \\ \midrule
Turk     &  $488,332$ &  \multirow{2}{*}{$2,000$} & \multirow{2}{*}{$359$} \\
ASSET    &   (Wiki-Auto)  &                      &                        \\
AutoMeTS & $1,967$                    & $411$                    & $418$                  \\ \bottomrule
\end{tabular}%
\caption{Numbers of sentence pairs in train, validation, and test sets used in our experiments}
\label{tab:corpus_stats}
\end{table}

As Turk and ASSET miss a training set, we employed Wiki-Auto~\cite{jiang-etal-2020-neural} following the convention. 
All of these corpora originate from English Wikipedia. 
\Tref{tab:corpus_stats} lists the numbers of training, validation, and test sets used. 

\subsection{Evaluation Metrics}
As the primary evaluation metric to measure the quality of sentence simplification, we used \textbf{SARI}~\cite{xu-etal-2016-optimizing}. 
SARI evaluates the success rates of addition, keeping, and deletion of tokens by comparing model-generated simplifications with the source sentence and references. 
The final score is computed by averaging F1 scores of
addition, keep, and deletion operations. 
For fine-grained analysis of the effects of our edit-based constraints, we also report the individual scores of these operations. 

In addition, we also employed metrics commonly used in simplification, namely, \textbf{BLEU}~\cite{papineni-etal-2002-bleu},  Flesch-Kincaid Grade Level (\textbf{FKGL})~\cite{kincaid-1975}, and \textbf{BERTScore}~\cite{bertscore}. 
SARI, BLEU, and FKGL were computed by a standard evaluation package of EASSE~\cite{alva-manchego-etal-2019-easse}\footnote{The default $4$-gram BLEU and SARI were measured.} while BERTScore used its official implementation.\footnote{\url{https://github.com/Tiiiger/bert_score}}

\subsection{Baselines}
We compared our method to previous studies also using lexically constrained decoding, namely, \citet{kajiwara-2019-negative} and \citet{zetsu-etal-2022-lexically}. 
While lexically constrained decoding theoretically applies to any base models using decoders, we limited ourselves to a basic pretrained sequence-to-sequence model, namely BART~\cite{lewis-etal-2020-bart}, expecting to make the analysis simpler and easier. 
Exploration of the best possible performance by applying to superior base models, such as large language models (LLMs)~\cite{kew-etal-2023-bless} or models trained with large corpus~\cite{martin-etal-2022-muss} constitutes our future work. 
We also compare our method to \citet{agrawal-2021} as the strong previous study that takes the same hybrid approach to simplification with us (see \Secref{sec:related_work}). 
In addition, the performance of naive beam search on the fine-tuned BART-base was also examined as the bottom line.

\begin{table*}[t]
\centering
\begin{tabular}{@{}lcccccccc@{}}
\toprule
                                  & SARI ($\uparrow$) & add           & keep          & del           & BLEU ($\uparrow$) & FKGL ($\downarrow$) & BS ($\uparrow$) & Len  \\ \toprule
\multicolumn{9}{c}{Turk}                                                                                                                                                    \\\midrule
Beam search                       & $39.7$             & $4.6$           & $58.7$          & $55.8$          & $35.2$             & $6.5$                & $91.7$                  & $13.6$ \\
\citet{agrawal-2021}              & $42.0$             & $4.0$           & $63.9$          & $58.0$          & $27.9$             & $7.4$                & $90.4$                  & $27.1$ \\
\citet{kajiwara-2019-negative}    & $38.1$             & $1.7$           & $64.9$          & $47.7$          & $22.6$             & $\bf{5.3}$       & $89.5$                  & $36.9$ \\
\citet{zetsu-etal-2022-lexically} & $50.7$             & $10.8$          & $74.9$          & $66.4$          & $53.4$             & $7.2$                & $94.8$                  & $16.9$ \\
Proposed method                   & $\bf{55.4}$    & $\bf{21.0}$ & $\bf{76.5}$ & $\bf{68.7}$ & $\bf{56.9}$    & $7.6$                & $\bf{94.9}$         & $18.4$ \\\midrule
\multicolumn{9}{c}{ASSET}                                                                                                                                                   \\\midrule
Beam search                       & $38.4$             & $5.7$           & $53.0$          & $56.4$          & $30.7$             & $6.5$                & $\bf{93.9}$         & $13.6$ \\
\citet{agrawal-2021}              & $43.1$             & $3.6$           & $64.4$          & $61.4$          & $41.1$             & $7.3$                & $92.7$                  & $17.4$ \\
\citet{kajiwara-2019-negative}    & $37.8$             & $2.6$           & $62.2$          & $48.5$          & $21.5$             & $\bf{5.8}$       & $89.5$                  & $36.2$ \\
\citet{zetsu-etal-2022-lexically} & $48.3$             & $10.4$          & $65.8$          & $\bf{68.6}$ & $40.2$             & $6.1$                & $93.6$                  & $14.8$ \\
Proposed method                   & $\bf{50.8}$    & $\bf{17.0}$ & $\bf{66.8}$ & $\bf{68.6}$ & $\bf{43.7}$    & $6.8$                & $93.8$                  & $16.7$ \\\midrule
\multicolumn{9}{c}{AutoMeTS}                                                                                                                                                \\\midrule
Beam search                       & $35.9$             & $0.8$           & $57.5$          & $49.4$          & $43.9$             & $11.5$               & $91.5$                  & $27.1$ \\
\citet{agrawal-2021}              & $38.9$             & $0.6$           & $45.3$          & $70.8$          & $20.1$             & $\bf{6.4}$       & $86.7$                  & $20.5$ \\
\citet{kajiwara-2019-negative}    & $42.9$             & $3.2$           & $59.2$          & $66.3$          & $29.3$             & $10.7$               & $88.9$                  & $35.8$ \\
\citet{zetsu-etal-2022-lexically} & $49.3$             & $3.7$           & $69.4$          & $74.8$          & $44.2$             & $9.2$                & $92.0$                  & $22.3$ \\
Proposed method                   & $\bf{52.3}$    & $\bf{8.0}$  & $\bf{72.9}$ & $\bf{75.9}$ & $\bf{50.0}$    & $9.1$                & $\bf{92.3}$         & $25.2$ \\ \bottomrule
\end{tabular}%
\caption{Evaluation results with oracle constraints; the scores were measured on the single references from which the constraints were extracted (`BS' and `Len' represent BERTScore and average output length, respectively).}
\label{tab:results_oracle_singleref}
\end{table*}

\subsection{Implementation} 
The baseline methods were replicated using the codes released by the authors. 
For methods using lexically constrained decoding, we fine-tuned BART-base\footnote{\url{https://huggingface.co/facebook/bart-base}} with an AdamW~\cite{adamw} optimiser using the corresponding training set of each corpus. 
The fine-tuning was conducted for $5$ epochs at maximum with early stopping based on validation SARI score. 
For inference, we set the beam size to $20$. 

The proposed method was implemented utilising the released codes of NeuroLogic Decoding\footnote{\url{https://github.com/GXimingLu/neurologic_decoding}}. 
Its hyper-parameters, namely, $\lambda^I$, $\lambda^D$, $\lambda^R$, and $\delta$, were searched using Optuna~\cite{optuna_2019} to maximise validation SARI score. 
The $\lambda$ values were searched in the range of $[0, 1]$, while the $\delta$ value was searched in $[0, 2]$\footnote{This is because when $\lambda$ ranges in $[0, 1]$, the maximum gap between edit scores is at most $2$ for each timestep.} with $100$ attempts. 
For $\alpha$, we used the default value due to its limited impact on the performance. 
\Secref{sec:hypara-setting} describes more details of implementation and shows hyper-parameter values.

\section{Evaluation with Oracle Constraints}
\label{sec:oracle_eval}
We first measure the performance of the proposed method under a setting where highly reliable constraints are given. 
This setting eliminates possible performance variations depending on the quality of constraints and clarifies how different lexically constrained decoding mechanisms contribute to sentence simplification.

\subsection{Oracle Constraint Extraction}
\label{sec:oracle_constraint_extraction}
We extracted highly reliable constraints from references. 
Specifically, we conducted word alignment between the source and reference sentences in a corpus using the state-of-the-art word aligner~\cite{lan-etal-2021-neural}. 
The alignment results were converted to constraints using simple heuristics below, which are referred to as `oracle' constraints hereafter\footnote{To be precise, these constraints are imperfect due to alignment errors; we assume their effects are minimal in this study.}. 
{
\setlength{\leftmargini}{15pt}         
\begin{itemize}
\setlength{\itemsep}{0pt}      
    \item Source tokens of null alignment were set as deletion operations (the proposed method) or negative constraints (baselines). 
    \item Source tokens aligned to the identical tokens were set as insertion operations (the proposed method)  or positive constraints (baselines).\footnote{We discarded reference tokens of null alignment from insertion constraints assuming that it is challenging in practice to predict new content to add based on a given sentence.} 
    \item Source tokens aligned to ones with different surfaces were regarded as substitution operations (the proposed method) or a set of negative and positive constraints (baselines). 
\end{itemize}
}

While Turk and ASSET provide multiple references, 
we used the every first reference of each source sentence for extracting the oracle constraints to keep the constraints and reference consistent.  
Therefore, the evaluation metrics were computed with these first references to meet the condition of oracle constraint extraction. 

\begin{table*}[t]
\centering
\begin{tabular}{cc}
\begin{minipage}[c]{0.48\textwidth}
\centering
\resizebox{\linewidth}{!}{
\begin{tabular}{@{}lcccccc@{}}
\toprule
             & \multicolumn{2}{c}{Turk}                           & \multicolumn{2}{c}{ASSET}                          & \multicolumn{2}{c}{AutoMeTS}                       \\
             & Comp. & Ours           & Comp. & Ours           & Comp. & Ours           \\ \midrule
Insertion    & $\bf{92.2}$                    & $91.9$          & $96.6$                             & $\bf{97.1}$ & $95.4$                             & $\bf{96.3}$ \\
Deletion     & $82.0$                             & $\bf{89.9}$ & $89.0$                             & $\bf{94.0}$ & $86.3$                             & $\bf{93.1}$ \\
Substitution & $17.9$                             & $\bf{55.3}$ & $20.9$                             & $\bf{53.1}$ & $10.1$                             & $\bf{44.3}$ \\ \bottomrule
\end{tabular}
}
\caption{Percentage of satisfied constraints; `Comp.' represents~\cite{zetsu-etal-2022-lexically}.}
\label{tab:tab:constraint-satisfaction-rate}
\end{minipage}
&
\begin{minipage}[c]{0.48\textwidth}
\centering
\resizebox{\linewidth}{!}{%
\begin{tabular}{@{}p{0.8\linewidth}rr@{}}
\toprule
                                                                               & \multicolumn{1}{c}{Comp.} & \multicolumn{1}{c}{Ours} \\ \midrule
Failed to generate named entities                                              & $6$                            & $3$                                                     \\
Dropped information when a sentence is complex and has multiple clauses & $15$                            & $4$                                                    \\
Altered meaning or focus of the sentence                            & $4$                            & $3$                                                     \\ \bottomrule
\end{tabular}%
}
\caption{Results of manual error analysis; `Comp.' represents~\cite{zetsu-etal-2022-lexically}}
\label{tab:manual_error_analysis}
\end{minipage}
\end{tabular}
\end{table*}

\begin{table*}[t]
\centering
\begin{tabular}{@{}lp{0.75\linewidth}@{}}
\toprule
src                               & He left a detachment of 11,000 troops to garrison the newly conquered region .                                      \\
reference                         & He left 11,000 troops there to defend the region .                                                                  \\ \midrule
\citet{agrawal-2021}              & He left a of 11 000 troops to the conquered region .                                                                \\
\citet{zetsu-etal-2022-lexically} & He left 11,000 troops .                                                                                             \\
Proposed method                   & He left 11,000 troops to defend the region .                                                                        \\
~~~edit-constraints                  & [insert] \textbf{left}, \textbf{troops} [delete] \textbf{detachment}, \textbf{newly} [substitute] \textbf{(garrison, defend)}                                    \\ \bottomrule
src                               & by most accounts , the instrument was nearly impossible to control .                                                \\
reference                         & most people said that the device was very hard to control .                                                         \\ \midrule
\citet{agrawal-2021}              & the instrument was to control .                                                                                     \\
\citet{zetsu-etal-2022-lexically} & It was almost impossible to control .                                                                               \\
Proposed method                   & Most people did not know how to control it , and most of them thought it was very hard .                            \\
~~~edit-constraints                  & [insert] \textbf{most}, was, \textbf{control} [substitute] (accounts, said), (instrument, device), \textbf{(nearly, very)}, \textbf{(impossible, hard)} \\ \bottomrule
\end{tabular}%
\caption{Example outputs of simplification models (\textbf{bold} constraints are satisfied by the proposed method.)}
\label{tab:examples}
\end{table*}

\subsection{Results}
\Tref{tab:results_oracle_singleref} shows results on the test sets under the oracle setting. 
The proposed method largely outperforms the previous studies. 
Breakdown scores of SARI indicate that our method improved all operations. 
We conjecture that this is because the stricter satisfaction conditions allowed proper handling of edit-based constraints. 

Among the methods employing lexically constrained decoding, \citet{kajiwara-2019-negative} uses only negative constraints. 
\Tref{tab:results_oracle_singleref} indicates that this method generated significantly longer sentences than others. 
As we observed their outputs, sentences were often disfluent with repeated tokens. 
We conjecture that this is an adverse effect using multiple negative constraints, which may have corrupted the decoder.

\begin{table*}[t]
\centering
\begin{tabular}{@{}lcccccccc@{}}
\toprule
                                  & SARI ($\uparrow$) & add          & keep          & del           & BLEU ($\uparrow$) & FKGL ($\downarrow$) & BS ($\uparrow$) & Len  \\ \toprule
\multicolumn{9}{c}{Turk}                                                                                                                                            \\ \midrule
Beam search                       & $39.7$             & $4.6$          & $58.7$          & $55.8$          & $35.2$             & $6.5$                & $91.7$           & $13.6$ \\
\citet{agrawal-2021}              & $38.6$             & $2.9$          & $58.9$          & $53.8$          & $37.0$             & $6.6$                & $90.7$           & $16.7$ \\
\citet{kajiwara-2019-negative}    & $33.4$             & $2.9$          & $\bf{64.8}$ & $32.5$          & $\bf{49.5}$    & $8.7$                & $\bf{92.9}$  & $19.7$ \\
\citet{zetsu-etal-2022-lexically} & $42.2$             & $\bf{8.5}$ & $59.2$          & $\bf{58.9}$ & $36.0$             & $\bf{6.0}$       & $91.9$           & $14.3$ \\
Proposed method                   & $\bf{42.6}$    & $7.9$          & $61.5$          & $58.5$          & $38.9$             & $6.4$                & $92.1$           & $15.3$ \\ \midrule
\multicolumn{9}{c}{ASSET}                                                                                                                                           \\ \midrule
Beam search                       & $38.4$             & $5.7$          & $53.0$          & $56.4$          & $30.7$             & $6.5$                & $\bf{93.9}$  & $13.6$ \\
\citet{agrawal-2021}              & $36.7$             & $2.2$          & $53.3$          & $54.6$          & $31.5$             & $6.6$                & $91.5$           & $16.7$ \\
\citet{kajiwara-2019-negative}    & $32.8$             & $2.3$          & $\bf{60.8}$ & $35.2$          & $\bf{44.0}$    & $8.7$                & $\bf{93.9}$  & $19.7$ \\
\citet{zetsu-etal-2022-lexically} & $\bf{40.9}$    & $\bf{7.5}$ & $54.8$          & $\bf{60.4}$ & $31.7$             & $\bf{6.0}$       & $92.7$           & $14.3$ \\
Proposed method                   & $\bf{40.9}$    & $7.4$          & $56.2$          & $59.1$          & $33.9$             & $6.4$                & $92.9$           & $15.3$ \\ \midrule
\multicolumn{9}{c}{AutoMeTS}                                                                                                                                        \\ \midrule
Beam search                       & $35.9$             & $0.8$          & $57.5$          & $49.4$          & $43.9$             & $11.5$               & $91.5$           & $27.1$ \\
\citet{agrawal-2021}              & $36.0$             & $0.4$          & $38.9$          & $\bf{68.7}$ & $16.1$             & $\bf{6.2}$       & $86.0$           & $19.4$ \\
\citet{kajiwara-2019-negative}    & $42.2$             & $\bf{6.5}$ & $\bf{63.4}$ & $56.8$          & $\bf{44.7}$    & $11.6$               & $\bf{92.4}$  & $31.1$ \\
\citet{zetsu-etal-2022-lexically} & $41.6$             & $2.0$          & $56.8$          & $66.0$          & $34.3$             & $8.6$                & $90.7$           & $21.3$ \\
Proposed method                   & $\bf{42.9}$    & $2.4$          & $60.2$          & $66.2$          & $38.9$             & $9.0$                & $91.2$           & $23.7$ \\ \bottomrule
\end{tabular}%
\caption{Evaluation results with predicted constraints; the scores were measured on the same single references with \Tref{tab:results_oracle_singleref} (`BS' and `Len' represent BERTScore and average output length, respectively).}
\label{tab:results_pred_singleref}
\end{table*}

In contrast, \citet{zetsu-etal-2022-lexically} consider both positive and negative constraints, which may enable it to avoid this corruption problem. 
As we discussed in \Secref{sec:edit-based-constraints}, their conditions of constraint satisfaction are looser than ours. 
\Tref{tab:tab:constraint-satisfaction-rate} shows the percentages of constraints satisfied by \citet{zetsu-etal-2022-lexically} and the proposed method, where our method achieved significantly higher constraint satisfaction rates. 
We further investigated the differences between \citet{zetsu-etal-2022-lexically} and our method by error analysis of their simplification outputs. 
We sampled $100$ source sentences from the ASSET corpus and manually annotated errors to simplification of each method. 
\Tref{tab:manual_error_analysis} shows the three major error types and their numbers of occurrences. 
The proposed method consistently decreased errors of all types, in particular, avoided overly dropping information when a source sentence is syntactically complex. 
These results indicate that our strict conditions for edit-constraint satisfaction effectively avoid sub-optimal generation. 

\Tref{tab:examples} shows example outputs.\footnote{We omitted outputs of \citet{kajiwara-2019-negative} due to their disfluent generation.} 
For both cases, \citet{agrawal-2021} rewrote less, which has been known as the conservativeness problem of the simplification models (see \Secref{sec:related_work}). 
In contrast, \citet{zetsu-etal-2022-lexically} and our method addressed this problem. 
Indeed, \citet{agrawal-2021} generated exactly the same sentences as their sources for $10.0\%$ cases on ASSET, while the proposed method decreased it down to $4.5\%$. 
In addition, the proposed method satisfied more constraints compared to \citet{zetsu-etal-2022-lexically} by avoiding sub-optimal generation.

\section{Evaluation with Predicted Constraints}
\label{sec:practical_eval}

In this section, we evaluate the proposed method under a practical setting, where constraints are predicted by a simple neural model. 

\subsection{Constraint Prediction}
We employed the simple constraint prediction model developed by \citet{zetsu-etal-2022-lexically}. 
It first predicts which tokens in a source sentence should be inserted, deleted, and substituted as a token classification problem. 
Specifically, we fine-tuned BERT-base~\cite{devlin-etal-2019-bert}\footnote{\url{https://huggingface.co/google-bert/bert-base-uncased}} using the oracle constraints extracted in \Secref{sec:oracle_constraint_extraction} as described in the original paper.

For tokens predicted as substitution, replacing tokens were determined using a lexical translation table, which was assembled from each training set using the Moses toolkit~\cite{koehn-etal-2007-moses}
.\footnote{While \citet{zetsu-etal-2022-lexically} fine-tuned a pretrained model for predicting replacing tokens, our preliminary experiment confirmed our simpler method utilising the translation table is superior and computationally faster. Note that we reused the same word alignment results computed for creating the oracle constraints for assembling the translation table. Furthermore, we postprocessed the translation table to discard target tokens with probabilities smaller than $0.002$.} 
The pairs of tokens were set as substitution operations in the proposed method (or a set of negative and positive constraints in baselines). 
When a token had multiple substitution candidates, these pairs were handled as `OR' constraints.

\subsection{Results}
\label{sec:results_predicted_constraints}

\Tref{tab:results_pred_singleref} shows results on test sets where constraints were predicted.\footnote{The simplification outputs on Turk and ASSET become equivalent because their source sentences and thus corresponding predicted constraints are the same, as can be seen in the equal values in FKGL and Len.} 
To ensure side-by-side comparison with \Tref{tab:results_oracle_singleref}, the evaluation metrics were computed using the same single references on Turk and ASSET (i.e., every first reference). 
\Secref{sec:more_results} provides the results computed on multiple references, which show the same trends with \Tref{tab:results_pred_singleref}. 
While the gains became smaller, the proposed method consistently outperformed the baselines on SARI. 
The results of \citet{agrawal-2021} and \citet{kajiwara-2019-negative} revealed their instability; they can be comparable to or even worse than the naive beam search, depending on the corpus. 
In contrast, our method achieved consistent improvements over the baselines. 

\begin{table}[t]
\centering
\begin{tabular}{@{}lccc@{}}
\toprule
                                                      & Turk  & ASSET & AutoMeTS \\ \midrule
Insertion: $C^I$                                      & $80.5$ & $74.6$ & $76.6$    \\
Deletion: $C^D$                                       & $50.7$ & $54.9$ & $66.1$    \\
Substitution: $C^R$                                   & $23.2$ & $23.8$ & $43.3$    \\
\multicolumn{1}{c}{~~from: $C^{R_i}$}                         & $29.2$ & $28.1$ & $43.3$    \\
\multicolumn{1}{c}{~~to: $C^{R_o}$} & $56.8$ & $62.2$ & $73.3$    \\ \bottomrule
\end{tabular}
\caption{Precision of predicted constraints (\%)}
\label{tab:precision-constraints}
\end{table}

Obviously, the quality of simplification by these methods depends on the quality of constraint prediction. 
\Tref{tab:precision-constraints} shows the precision of each edit constraint compared to the oracle constraints. 
As expected from the naive design of the prediction model, the results indicate that there is a large room for improvement in constraint prediction, particularly in deletion and substitution operations. 
The development of a sophisticated prediction model constitutes our future work. 
Nonetheless, it is remarkable that the proposed method outperforms the previous studies even with imperfect constraints. 

\section{Summary and Future Work}
In this study, we proposed the edit-constrained decoding for sentence simplification. 
Our constraints directly replicate the edit operations involved in simplification and are associated with strict satisfaction conditions. 
These features allow us to avoid the sub-optimal generation observed in previous studies. 
The evaluation with oracle constraints confirmed that our method largely outperforms the previous methods using lexically constrained decoding. 
Furthermore, the evaluation with a simple constraint prediction model confirmed consistent improvement by our method over the previous studies even with imperfect constraints. 

In future work, we will apply our edit-constrained decoding to LLMs to further enhance the sentence simplification technology. 
As \citet{valentini-etal-2023-automatic} revealed, LLMs still lack the ability to control lexical complexities. 
The proposed method is promising as complementation on this issue. 
We will also improve the constraint prediction model, which should boost the quality of simplification by a large margin. 

\section*{Limitations}
To derive the full potential of our method for sentence simplification, we should improve the quality of constraint prediction as discussed in \Secref{sec:results_predicted_constraints}. 
Employment of LLMs is promising given their high performances in simplification~\cite{kew-etal-2023-bless}. 

We also observed that there is room for improvement in the mechanisms of lexically constrained decoding itself. 
While existing methods use constant weights for constraints and attempt to apply them since the beginning of generation, this often excessively affects generation and ends up in degenerate solutions. 
Considering the generation likelihood changes over timesteps, the constraint weights should be adjusted depending on the generation states. 
This direction also constitutes our future work. 

\section*{Acknowledgments}
This work was supported by Cross-ministerial Strategic Innovation Promotion Program (SIP) on “Integrated Health Care System” Grant Number JPJ012425 and JSPS KAKENHI Grant Number JP21H03564.

\bibliography{arase}

\appendix
\section{Additional Experiment Details}

\subsection{Implementation Details}
\label{sec:hypara-setting}

All the experiments were conducted on NVIDIA RTX A6000 (48GB memory) GPUs installed on a Ubuntu server. 
The main memory size was 1TB, and the central processor was AMD EPYC CPU.

\begin{table}[hbt!]
\centering
\resizebox{\linewidth}{!}{%
\begin{tabular}{@{}lcccccc@{}}
\toprule
            & \multicolumn{2}{c}{Turk} & \multicolumn{2}{c}{ASSET} & \multicolumn{2}{c}{AutoMeTS} \\ 
            & oracle    & predicted    & oracle     & predicted    & oracle      & predicted      \\ \midrule
$\lambda^I$ & $0.11$      & $0.01$         & $0.05$       & $0.01$         & $0.03$        & $0.05$           \\
$\lambda^D$ & $0.66$      & $0.94$         & $0.67$       & $0.94$         & $0.81$        & $0.94$           \\
$\lambda^R$ & $0.23$      & $0.05$         & $0.28$       & $0.05$         & $0.16$        & $0.01$           \\
$\delta$    & $0.12$      & $0.18$         & $0.14$       & $0.18$         & $0.10$         & $0.10$            \\ \bottomrule
\end{tabular}%
}
\caption{Hyper-parameter settings }
\label{tab:hypara-setting}
\end{table}

\begin{table*}[hbt!]
\centering
\begin{tabular}{@{}lcccccccc@{}}
\toprule
                                  & SARI ($\uparrow$) & add          & keep          & del           & BLEU ($\uparrow$) & FKGL ($\downarrow$) & BS ($\uparrow$) & Len  \\ \toprule
\multicolumn{9}{c}{Turk}                                                                                                                                                \\ \midrule
Beam search                       & $36.3$              & $2.8$          & $60.6$          & $45.6$          & $71.3$              & $6.5$                 & $93.9$            & $13.6$ \\
\citet{agrawal-2021}              & $37.3$              & $1.7$          & $62.8$          & $47.4$          & $68.2$              & $6.6$                 & $92.7$            & $16.7$ \\
\citet{kajiwara-2019-negative}    & $36.6$              & $2.6$          & $\bf{72.9}$ & $34.3$          & $\bf{89.9}$     & $8.7$                 & $\bf{95.8}$   & $19.7$ \\
\citet{zetsu-etal-2022-lexically} & $39.2$              & $4.9$          & $61.4$          & $\bf{51.2}$ & $69.1$              & $\bf{6.0}$        & $93.8$            & $14.3$ \\
Proposed method                   & $\bf{40.0}$     & $\bf{5.2}$ & $63.8$          & $50.9$          & $72.0$              & $6.4$                 & $94.1$            & $15.3$ \\ \midrule
\multicolumn{9}{c}{ASSET}                                                                                                                                               \\ \midrule
Beam search                       & $38.9$              & $3.2$          & $54.7$          & $58.6$          & $84.3$              & $6.5$                 & $96.0$            & $13.6$ \\
\citet{agrawal-2021}              & $38.5$              & $1.6$          & $55.7$          & $58.1$          & $71.7$              & $6.6$                 & $94.4$            & $16.7$ \\
\citet{kajiwara-2019-negative}    & $33.4$              & $2.2$          & $\bf{60.1}$ & $38.0$          & $\bf{87.8}$     & $8.7$                 & $\bf{97.5}$   & $19.7$ \\
\citet{zetsu-etal-2022-lexically} & $42.1$              & $5.3$          & $57.2$          & $\bf{63.9}$ & $80.6$              & $\bf{6.0}$        & $96.1$            & $14.3$ \\
Proposed method                   & $\bf{42.2}$     & $\bf{5.6}$ & $58.4$          & $62.5$          & $82.4$              & $6.4$                 & $96.3$            & $15.3$ \\ \bottomrule
\end{tabular}%
\caption{Evaluation results with predicted constraints; the scores were measured using all of the multi-references (`BS' and `Len' represent BERTScore and average output length, respectively).}
\label{tab:results_pred_multiref}
\end{table*}

We used a fine-tuned BART-base as the sequence-to-sequence generation model. 
For experiments targeting ASSET, we reused the model fine-tuned employing Turk's validation set (for early stopping) to save computational costs. 
Given that the source sentences of Turk and ASSET are common, we assume this setting does not significantly deteriorate the model's performance on ASSET.

\Tref{tab:hypara-setting} shows the hyper-parameter values on the proposed method. 
To speed up the hyper-parameter search, we used $300$ random samples from the validation sets on Optuna. 
When evaluating the performance on Turk and ASSET with predicted constraints, we used the common hyper-parameter values tuned on Turk to save computational costs. 

The hyper-parameter values in \Tref{tab:hypara-setting} are considerably different between the oracle and predicted constraints. 
We conjecture the discrepancy is due to the combination of the generation model performance and constraint prediction quality. 
The plain beam search on fine-tuned BART (`Beam search' rows in \Tref{tab:results_oracle_singleref} and \Tref{tab:results_pred_singleref}) has a much higher deletion score than the addition score (see `del' and `add' columns), which indicates that the generation model is originally weaker on insertion and substitution operations. 
As oracle constraints are completely reliable, our method can put more emphasis on them to promote these edit operations. 
In contrast, predicted constraints are imperfect as shown in \Tref{tab:precision-constraints}, our method cannot increase weights on these operations, which may deteriorate their performance.

\subsection{More Experiment Results}
\label{sec:more_results}

\Tref{tab:results_pred_multiref} shows results when constraints were predicted, where the scores were measured on test sets of Turk, ASSET, and AutoMeTS using all of the multi-references. 
The trend is the same with \Tref{tab:results_pred_singleref}; the proposed method consistently outperforms the baselines. 

\end{document}

%% file: math_commands.tex
\newcommand\identity{1\kern-0.25em\text{l}}

\def\Figref#1{Figure~\ref{#1}}

\def\Secref#1{\S\ref{#1}}

\def\eqref#1{equation~(\ref{#1})}
\def\Eqref#1{Equation~(\ref{#1})}

\def\1{\bm{1}}

\DeclareMathAlphabet{\mathsfit}{\encodingdefault}{\sfdefault}{m}{sl}
\SetMathAlphabet{\mathsfit}{bold}{\encodingdefault}{\sfdefault}{bx}{n}

\DeclareMathOperator*{\argmax}{arg\,max}